\documentclass[pmlr]{jmlr} % new name PMLR (Proceedings of Machine Learning Research)

\usepackage[T1]{fontenc}

\usepackage{booktabs}
\usepackage{graphicx}
\usepackage{microtype}
\usepackage{paralist}
\usepackage{enumitem}

\newcommand{\tagdsmode}{proceedings}

\makeatletter
\newcommand{\tagdssubmission}{submission}
\newcommand{\tagdsproceedings}{proceedings}

\ifx\tagdsmode\tagdsproceedings
\else\ifx\tagdsmode\tagdssubmission
  \def\ps@jmlrtps{%
    \let\@mkboth\@gobbletwo
    \def\@oddhead{\scriptsize Under Review at the 2nd Conference on Topology, Algebra, and Geometry in Data Science\hfill}%
    \let\@evenhead\@oddhead
    \def\@oddfoot{}%
    \let\@evenfoot\@oddfoot
  }

\else
  \def\ps@jmlrtps{%
    \let\@mkboth\@gobbletwo
    \def\@oddhead{}%
    \let\@evenhead\@oddhead
    \def\@oddfoot{}%
    \let\@evenfoot\@oddfoot
  }
\fi\fi
\makeatother

\usepackage{longtable}% for long tables

\usepackage{booktabs}
\usepackage[load-configurations=version-1]{siunitx} % newer version
\theorembodyfont{\upshape}
\theoremheaderfont{\scshape}
\theorempostheader{:}
\theoremsep{\newline}

\jmlrvolume{334}
\jmlryear{2026}
\jmlrworkshop{Topology, Algebra, and Geometry in Data Science}

\title[Template]{TAG-DS Paper Template}

\ifx\tagdsmode\tagdssubmission

\else
 \author{\Name{Leon Dahlmeier} \Email{dleon@ethz.ch}\\
 \addr  Department of Mathematics, ETH Zürich, Zürich, Switzerland
 \AND
 \Name{Sara Kališnik} \Email{skalisnik@psu.edu}\\
 \addr Department of Mathematics, PSU, Pennsylvania, United States
 \AND
 \Name{Albert Mehl} \Email{albert.mehl@zzm.uzh.ch}\\
 \addr Department of Computerized Dentistry, Clinic of Conservative and Preventive Dentistry, University of Zürich, Zürich, Switzerland
 \AND
 \Name{Bastian Rieck} \Email{bastian.grossenbacher-rieck@unifr.ch}\\
 \addr Department of Computer Science, University of Fribourg, Fribourg, Switzerland
}

\fi

\usepackage{amssymb}
\usepackage{amsmath}

\usepackage{comment}
\usepackage{tikz}
\usepackage{tikz-cd}

\usepackage{diagbox}
\usepackage{makecell}  % For splitting content into multiple lines in table cells
\usepackage{slashbox}  % For diagonal cells like \backslashbox
\usepackage{colortbl}
\usepackage{xcolor}
\usepackage{soul}
\usepackage[labelfont=bf]{caption}
\usepackage{subcaption} 
\usepackage{multirow} 
\usepackage{rotating}

\usepackage{makecell} % 

\begin{document}
\title{Topology of a Smile: Persistent Homology in Dental Imaging}

\maketitle              % typeset the header of the contribution
\begin{abstract}
%The abstract should briefly summarize the contents of the paper in 150--250 words.
CBCT (Cone Beam Computed Tomography) scans provide detailed three-dimensional images, widely used in dentistry for diagnostic and treatment planning tasks. While invaluable, analyzing and documenting these scans is labor-intensive, prompting efforts to automate key steps like the classification and segmentation of anatomical structures to identify tooth types and associated pathologies.  In this article, we propose an approach to automation that leverages persistent homology, a framework from topological data analysis that studies the shape of data by identifying features like connected components, holes, and voids across multiple scales. Persistent homology, together with a support vector machine, allows us to classify teeth in a CBCT scan and to perform diagnostics. 
Our method advances the state of the art, reaching average accuracy scores of 97.67\% for tooth-labeling and 96.77\% for diagnostic tasks, outperforming a CNN trained on the same data with accuracy of 70.27\% and 86.67\%, respectively.
% For the tooth-labeling task, we used 555 teeth and reached scores ranging from 97.31\% to 98.55\% averaging at 97.89\%. For performing diagnostics, we used 600 teeth with a total of 10 different labels, reaching scores from 90.96\% to 99.13\% averaging at 96.77\%. 

% \keywords{Dental Cone Beam Computed Tomography  \and Topological Data Analysis \and Machine Learning \and Tooth Identification \and Dental Diagnostics.}
\end{abstract}
\begin{keywords}
Dental Cone Beam Computed Tomography, Topological Data Analysis, Machine Learning, Tooth Identification, Dental Diagnostics.
\end{keywords}

\section{Introduction}

Three-dimensional X-ray images (Cone Beam Computed Tomography, CBCT) are used, for example, before implant and orthodontic treatments, prior to the extraction of impacted and fractured teeth, and for classifying cystic lesions and other pathological findings, among other applications \citep{Lee2020,Morgan2023}. Additionally, the documentation can be used to monitor changes over time, thereby predicting the progression of diseases or controlling treatments \citep{Morgan2023}. CBCT data incorporate a wealth of information; however, for dentists, the 3D visualization is laborious and time-consuming; hence, there are various approaches to automatically extract relevant findings and generate documentation. Key steps in this workflow are the segmentation and the classification of anatomical structures \citep{Fontenele2023,Xiang2024}. In particular, it is crucial to identify the types of teeth present in the current CBCT and determine if they exhibit any specific alterations or pathological findings \citep{Xiang2024}.
Owing to the rise of deep neural networks, approaches for extracting information and pathological findings have been tested and implemented for a variety of radiographic images. 
In dentistry, the primary focus has traditionally been on 2D radiographs, such as panoramic images and bitewings~\citep{Hamamci2023}. However, in recent years, 3D X-ray imaging has gained increasing attention. Attempts to solve the automatic labeling task in CBCT data were described by \citet{Mikia2017} and \citet{Gerhardt2022}.
More general approaches combine tooth \emph{labeling} with tooth \emph{segmentation}~\citep{Cui2019,Ezhov2019,Shaheen2021,Zhou2025}.
A typical setting is described by \citet{Cui2022}, which performs tooth classification and segmentation of teeth and bony anatomical structures, reaching an accuracy~(sensitivity) of up to 93\%.
The common denominator of all these works is their use of convolutional architectures, making use of CNNs or more complex models like UNet~\citep{Polizzi2023}.
This makes the approaches infeasible to train or apply in data-scarce scenarios.

In this article, we introduce an automatic classification procedure using tools from topology, a branch of mathematics that studies the shape of objects. In particular, we employ persistent homology, an invariant that captures topological features across scales and has been shown to be effective for image-based analysis and classification~\citep{ADCOCK201436,carlsson2014topological,Edelsbrunner2002,LEVENSON2024102060,Vandaele23a}. We combine these topological descriptors with support vector machines to create a novel method for classifying teeth and pathological findings in CBCT data.
Unlike existing techniques, our method
\begin{inparaenum}[(i)]
  \item requires far fewer features, while being able to handle regions of interest~(ROIs) of \emph{varying size},
  \item is provably stable~\citep{Cohen-Steiner2007,kim2021noise}, and
  \item captures morphological features~\citep{liu2023phgnet,perez2023tda}.
\end{inparaenum}%The results of our method are compared to an existing CNN approach; we reach accuracy scores of 97.67\% for tooth-labeling and 96.77\% for diagnostics tasks, outperforming a CNN trained on the same data with accuracy of 70.27\% and 86.67\% respectively..

The results of our method are compared with an existing CNN-based approach \citep{ESMAEILYFARD2024328}, achieving 
accuracy scores of 97.67\% for tooth labeling and 96.77\% for diagnostic tasks, 
thereby outperforming a CNN trained on the same data, which attains accuracies of 
70.27\% and 86.67\%, respectively.

\section{Methods}
\subsection{Topological Persistence}

This section provides a brief and informal introduction to cubical complexes and persistent homology. For an in-depth treatment of the subject, see \citet{carlsson2014topological} or \citet{Edelsbrunner2002}.

\subsubsection{Cubical Complexes}\label{ref:cubical_complexes}

Our data, which consists of CBCT scans (as seen in Figure \ref{fig:CBCT-scan}), is given in the form of a stack of grayscale images.
To represent such images, we use \emph{cubical complexes}.
Informally, an $m$-dimensional cubical complex consists of $k$-dimensional elementary cubes (where $k\leq m$) glued together along common boundaries; $0$-dimensional, $1$-dimensional and $2$-dimensional elementary cubes are vertices, edges and squares, respectively. For the formal definition, please refer to \citet{CubicalComplexes}.  
%
%From an $n$-dimensional image, we construct a cubical complex by assigning an $n$-dimensional elementary cube to each $n$-pixel. We then include all lower-dimensional boundary cubes. For example, in the case of 2-dimensional images, we add zero-dimensional cubes (vertices) at each corner and one-dimensional cubes (edges) along each boundary of the pixels. To avoid redundancy, overlapping faces between neighboring pixels are represented by a single shared elementary cube. Figure~\ref{fig:complex} depicts the process of constructing a cubical complex from a grayscale image.

From an $n$-dimensional image, we construct a cubical complex by assigning an $n$-dimensional elementary cube to each $n$-pixel. We then include all lower-dimensional boundary cubes. %Figure~\ref{fig:complex} depicts the process of constructing a cubical complex from a grayscale image.

%For example, the cubical complex representing the image in Figure~\ref{fig:complex} shown on the right in Figure~\ref{fig:complex}, consists of nine 2-dimensional cubes (gray squares), 24 1-dimensional cubes (black lines), and 16 0-dimensional cubes (black discs, representing vertices).

%
%To incorporate the grayscale values into the cubical complex, instead of considering a \emph{single} cubical complex, we consider a one-parameter family of cubical complexes. This approach preserves the information encoded in the grayscale values that would otherwise be ``lost'' in a single-complex representation. 

To incorporate grayscale information, we consider a one-parameter family of cubical complexes rather than a single complex, thereby preserving information that would otherwise be lost in a single-complex representation.

\subsubsection{Cubical Filtrations}
\label{subsec:cubical_filtration}

\begin{figure}[h!]
    \centering
    \includegraphics[clip, trim=2cm 0.5cm 2cm 0.5cm, width=0.9\textwidth]{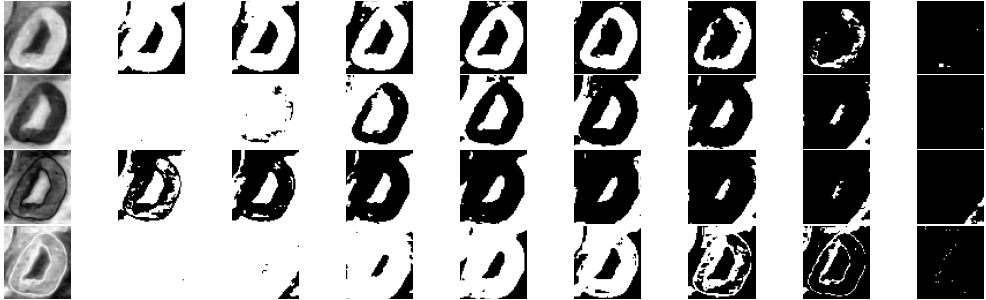}
    \caption{Sublevel filtration of the original, the inverted, the contrast, and the inverted contrast image, respectively.}
    \label{fig:filtration_17}
\end{figure}

A one-parameter family of cubical complexes together with inclusion maps is called a \emph{(cubical) filtration}~\citep{carlsson2014topological}.
Given an $n$-dimensional image and a function $f$ defined on the pixels (e.g., grayscale values), we construct a cubical complex $C_t$ by including an $n$-dimensional cube for each pixel $p$ with $f(p)\le t$, together with all boundary cubes. As $t$ increases, additional pixels are included until $C_t$ contains a cube for each pixel. This yields a nested sequence
\begin{equation}
\emptyset = C_0 \subseteq C_1 \subseteq C_2 \subseteq \cdots \subseteq C_n = C,
\end{equation}
referred to as a \emph{sublevel set filtration}.
In this paper, we consider the following images and filter functions for constructing cubical filtrations: the \emph{original image}, the \emph{ inverted image}, the \emph{ contrast image}, and the \emph{ inverted contrast} image (see the images on the left in Figure~\ref{fig:filtration_17}). The inverted image is generated by setting each pixel's  grayscale value $g_o$ to $g=g_{\max}-g_o$, where $g_{\max}$ denotes the highest grayscale value in the original image. The contrast image is generated by setting each pixel's old grayscale value $g_o$ to $h=\lvert g_o - 0.5\cdot(g_{\min} + g_{\max})\rvert $.
%\[
% g_{rev}=\min_{g \textrm{ is a grayscale value in the image}}  ( g + 0.5\cdot(g_{\max}-g_{\min})).\] 
For the inverted contrast image, we first take the contrast image of the original image, and then the inverted image of the contrast image. The choice of contrast image is inspired based on  considering carious lesions, as they appear darker than the neighboring healthy enamel or dentin but may still be brighter than surrounding soft tissue or air, making it harder for the lesion to appear as a whole in the sublevel set filtrations of the original scan.

To illustrate this cubical filtration procedure, consider Figure~\ref{fig:filtration_17}.
The top row shows the grayscale sublevel set filtration of the original image, while the remaining rows correspond to the inverted, contrast, and inverted contrast images. As the parameter increases in the top row, additional cubes are added according to pixel intensity, revealing topological features such as loops. The evolution of these features across the filtration is captured using persistent homology.

%To illustrate this procedure, consider the image in the top-left corner of Figure~\ref{fig:filtration_17}. 
%\begin{comment}
 %  \begin{figure}[h!]
 %   \centering
 %   \includegraphics[clip, trim=2cm 3cm 2cm 3cm, width=0.9\textwidth]{Images/basic_examples/filtration_of_17.pdf}
%    \caption{Sublevel filtration of the original and its inverted image.}
%    \label{fig:filtration_17}
%\end{figure} 
%\end{comment}
%The inverted image is generated by setting each pixel's old gray-scale-value $g_o$ to $g_i=g_{max}-g_o$, where $g_{max}$ denotes the highest gray-scale-value in the original image.

%The top row in Figure~\ref{fig:filtration_17} shows the grayscale sublevel set filtration consisting of 8 cubical complexes. The subsequent rows illustrate the filtrations of the inverted image, the contrast image, and the inverted contrast image. In the first complex of the top row, cubes appear only where pixels are entirely black. As the filtration progresses, additional cubes emerge, corresponding to increasingly brighter pixels, gradually revealing a loop within the image. 

%We capture the evolving shape of these spaces using persistent homology.
 %The filtration ends once even the white pixels get a representative. Persistent homology will now calculate birth and death times for each nontrivial generator of each homology group.
%${\displaystyle L_{c}^{-}(f)=\left\{(x_{1},\dots ,x_{n})\mid f(x_{1},\dots ,x_{n})\leq c\right\}}$

\subsubsection{Persistent Homology}

%Intuitively, the \( k \)-dimensional homology group of a cubical complex \( X \), denoted by \( \textbf{H}_k(X) \), captures information about the \( k \)-dimensional voids in \( X \). The dimension of \( \textbf{H}_k(X) \), known as the \( k \)-th Betti number \( \beta_k(X) \), provides a count of these voids. Specifically, \( \beta_0(X) \) represents the number of connected components, \( \beta_1(X) \) counts the loops, \( \beta_2(X) \) measures the number of voids, and more generally, \( \beta_k(X) \) counts the \( k \)-dimensional voids in \( X \). For example, a ring encloses a one-dimensional hole, resulting in \( \beta_1 = 1 \), while a sphere encloses a two-dimensional void, meaning that \( \beta_2 = 1 \). Both spaces consist of a single piece, so for both spaces \( \beta_0 = 1 \).

Homology is a topological invariant used to distinguish spaces by considering patterns within them. For a cubical complex $X$, the $k$-dimensional homology group $\mathbf{H}_k(X)$ encodes information about $k$-dimensional voids, and its dimension, the $k$-th Betti number $\beta_k(X)$, counts these features. In particular, $\beta_0(X)$ counts connected components, $\beta_1(X)$ counts loops, and $\beta_2(X)$ counts two-dimensional voids.%More generally, \( \beta_k(X) \) counts the \( k \)-dimensional voids in \( X \). 
%For example, a ring encloses a one-dimensional hole, resulting in \( \beta_1 = 1 \), while a sphere encloses a two-dimensional void, meaning that \( \beta_2 = 1 \). Since both spaces consist of a single piece, they have \( \beta_0 = 1 \).

\begin{figure}[h]
  \begin{minipage}[b]{0.3\textwidth}
    \includegraphics[clip, trim=3cm 0.5cm 3cm 0.5cm, width=0.9\textwidth]{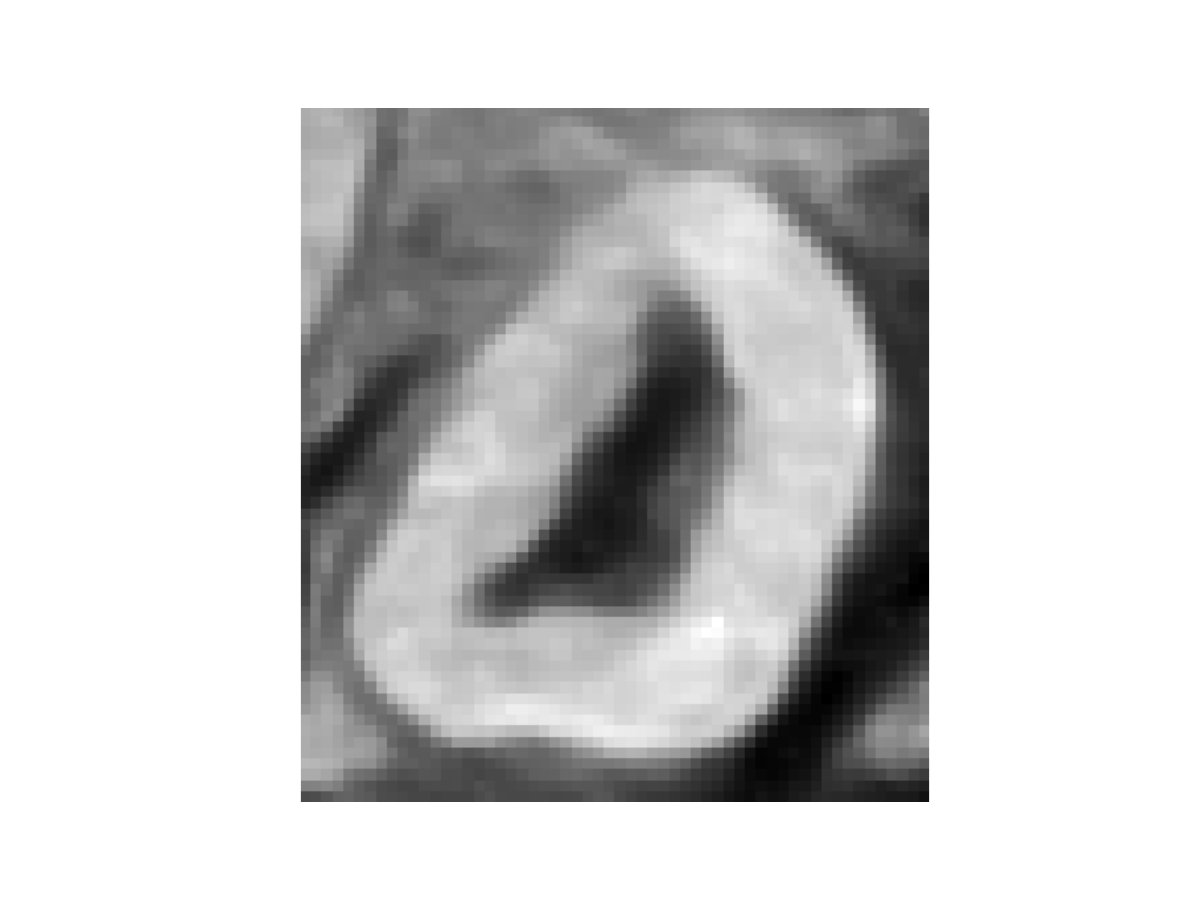}
    \caption{Image of a two-dimensional slice of a tooth, specifically 17.}
    \label{fig:slice_17}
      \hspace{0.2\linewidth}
  \end{minipage}
  \hfill
  \begin{minipage}[b]{0.3\textwidth}
    \includegraphics[width=\textwidth]{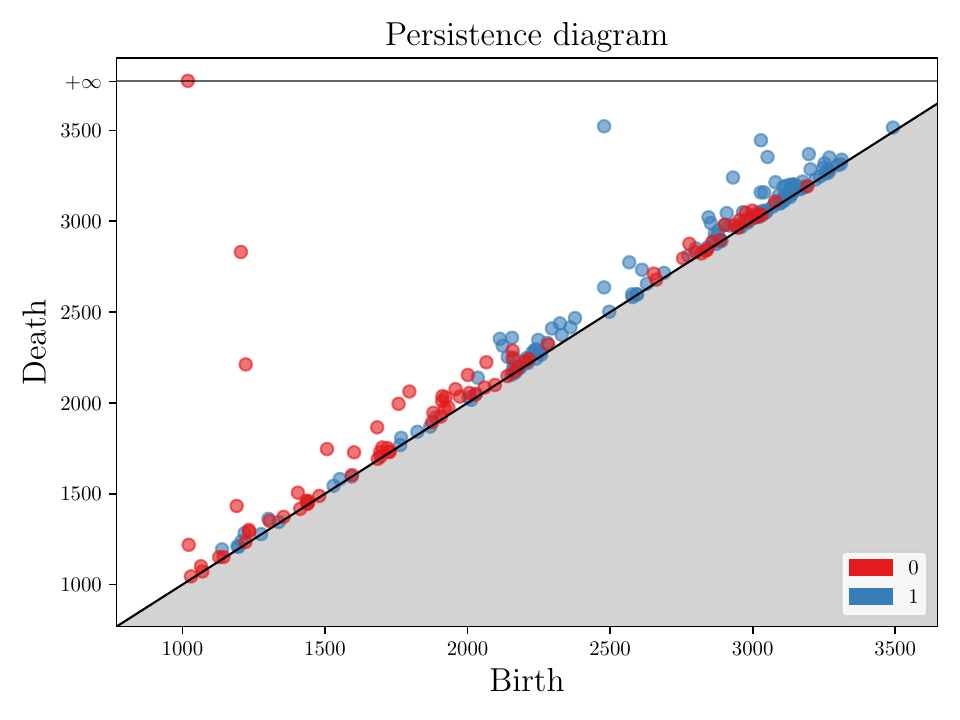}
    \caption{The persistence diagram for the sublevel set grayscale filtration of the image in Figure \ref{fig:slice_17}.}
    \label{fig:pdiag_of_17}
  \end{minipage}
  \hfill
  \begin{minipage}[b]{0.3\textwidth}
    \includegraphics[width=\textwidth]{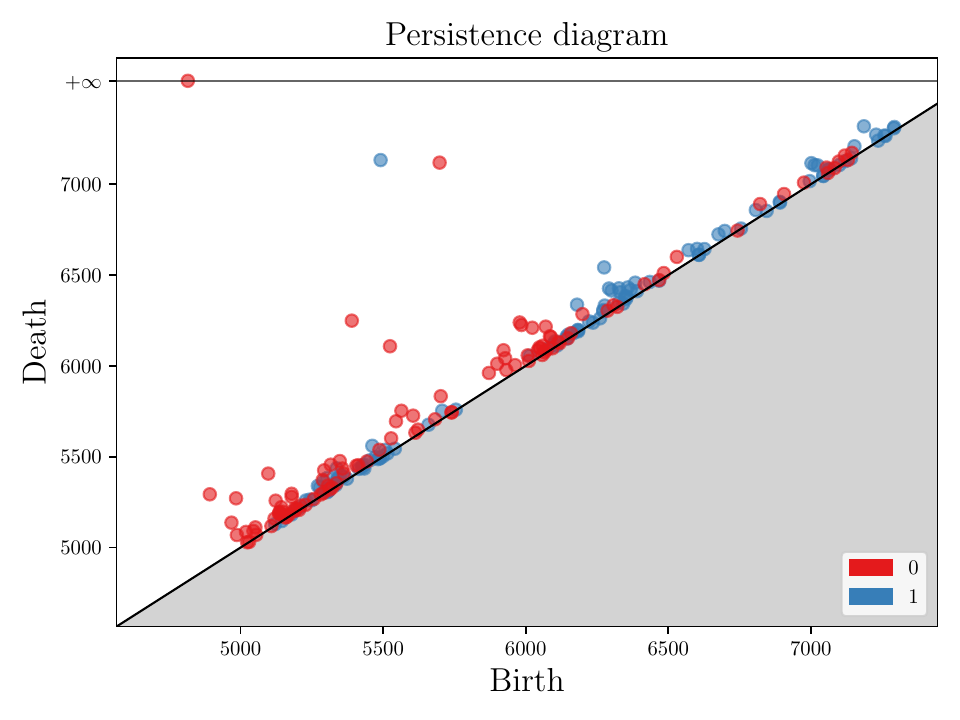}
    \caption{The persistence diagram for the sublevel set filtration of the inverted image of Figure~\ref{fig:slice_17}.}
    \label{fig:pdiag_of_inv_17}
  \end{minipage}
\end{figure}

Persistent homology is an adaptation of homology to the setting of filtrations. It allows us to track when topological features, such as connected components, loops and voids, appear and disappear in a filtration~\citep{carlsson2014topological,Edelsbrunner2002}. A \emph{persistence diagram} associated with a filtration is a collection of points \(\{(b_l, d_l)\}_{l \in L}\) in the extended plane \(\overline{\mathbb{R}}^2\), where \(\overline{\mathbb{R}} = \mathbb{R} \cup \{-\infty, \infty\}\). These points lie on or above the diagonal, with birth times \( b_l \) as the \( x \)-coordinates and death times \( d_l \) as the \( y \)-coordinates. The multiplicities of points on the diagonal are set to \( \infty \). The lifetime of the \(l\)-th feature is given by \(d_l - b_l\). 
Figures \ref{fig:pdiag_of_17} and \ref{fig:pdiag_of_inv_17} depict persistence diagrams arising from sublevel set filtrations for both the original image in Figure~\ref{fig:slice_17} and its inverted counterpart. In these diagrams, red points represent dimension-zero homology, while blue points indicate dimension-one homology.

%Points in the persistence diagram that are closer to the diagonal correspond to features with shorter lifetimes, which are more likely to arise from noise in the image. Conversely, points farther from the diagonal represent features with longer lifetimes, indicating they are more likely to reflect significant structures in the original image. Since our filtration always concludes with a fully filled-in image, there is always a connected component that persists to infinity.

%Looking at the complex in the top-right corner of Figure~\ref{fig:filtration_17}, we observe two holes that are not features of the tooth. However, since the gray-scale values of the corresponding pixels are similar to those of the surrounding pixels, these holes are quickly filled, resulting in short lifetimes. The blue dot farthest from the diagonal corresponds to the one-dimensional hole we expected. In Figure~\ref{fig:pdiag_of_17}, this hole is born when the outline of the tooth closes and dies when the tooth itself is included. Conversely, in Figure~\ref{fig:pdiag_of_inv_17}, the hole is born when the tooth is included and dies when its interior is fully filled.

\subsubsection{Vectorization Techniques}\label{Subsubsection:Vectorization Techniques}
While persistent homology is a powerful tool for extracting topological features from data, its output is non-Euclidean, making it difficult to integrate with standard machine learning methods that require vector inputs. Various vectorization techniques~\citep{JMLR:v18:16-337,Bub15,Kal18} address this by mapping persistence diagrams to fixed-dimensional representations while preserving topological information. For our dataset, \emph{persistent statistics}~\citep{Ali_2023} performed best, generating feature vectors from summary statistics of birth and death times (e.g., mean, standard deviation, median, and percentiles). Appendix \ref{The used vectorization method} compares this approach with several alternative vectorization methods and supports our choice.

%While persistent homology is a powerful tool for extracting topological features from data, its output is non-Euclidean, making it difficult to integrate with standard machine learning methods that require vector inputs. To address this challenge, various vectorization techniques~\citep{JMLR:v18:16-337,Bub15,Kal18} have been developed to transform persistence diagrams into fixed-dimensional representations while preserving key topological information. 

%For our dataset, \emph{persistent statistics}~\citep{vectorization} produced the best results. The feature vectors were generated by computing statistical measures, including but not limited to the mean, standard deviation, median, and percentiles of birth and death times.

\subsection{Data and Study Design}
\subsubsection{Data}

\begin{figure}
    \centering
    \includegraphics[width=0.7\linewidth, trim=5cm 8cm 5cm 8cm, clip]{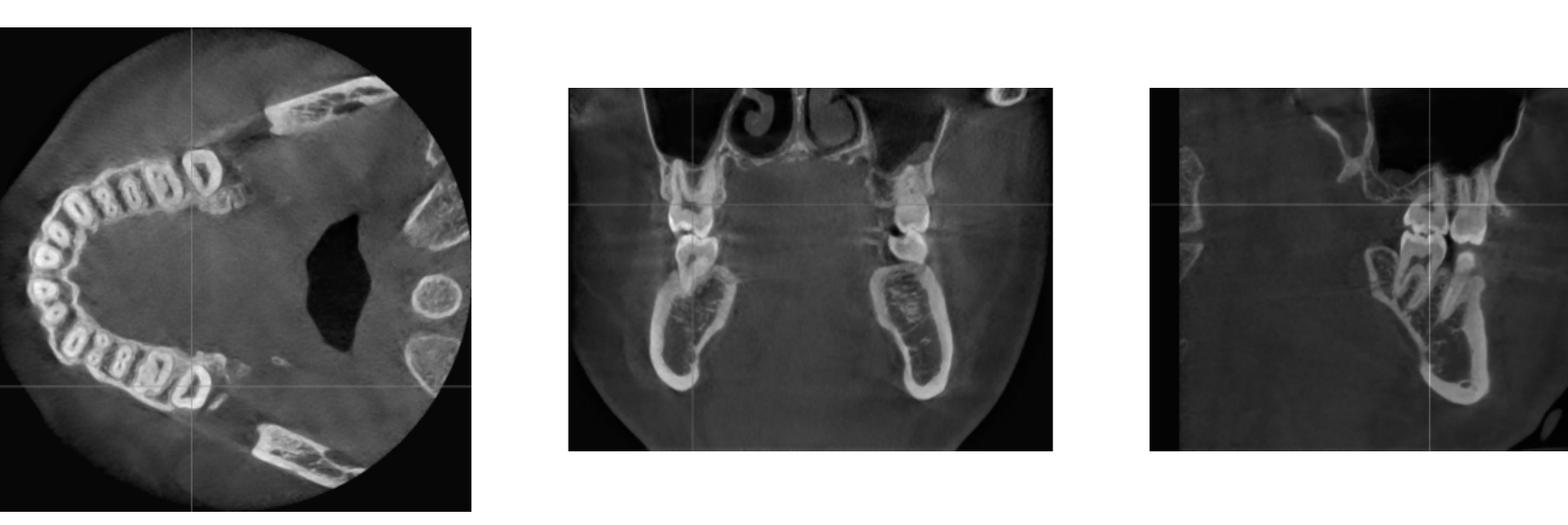}
    \caption{A CBCT scan represented by an axial, coronal and sagittal slice.}
    \label{fig:CBCT-scan}
\end{figure}

Twenty CBCT datasets, we have depicted one in Figure \ref{fig:CBCT-scan}, were randomly extracted from a larger data base of fully anonymized CBCT cases (n = 250). These cases were sourced from the X-ray archiving system of a university clinic, which includes patient cases used for examinations or surgical treatment planning as part of routine clinical practice (anonymized Ethics Committee). % BASEC Req-2021-01183, see Ref
 The inclusion criterion for the 20 CBCT datasets was the presence of a nearly full dentition $(\geq 24 \text{ teeth (excluding wisdom teeth))}$ and a dentation in occlusion.
 Each tooth was labeled with its FDI-label \citep{FDI-label} and associated diagnoses.
 This was performed by a dentist with more than three years of clinical experience. Our dataset contains diagnoses as follows. There are 51 fillings: 7 mesial, 42 occlusal, and 2 distal. Additionally, 44 teeth are held in place by a retainer. Among the 26 impacted teeth, 6 are covered only by gingiva, while 20 are also covered by bone. There are 7 cases of secondary caries and 3 cases of attrition. Finally, 476 teeth show no clinically relevant findings. In total, we use 600 teeth to perform diagnostics. For the FDI-labeling task, we exclude wisdom teeth, as too many of them are impacted. All other teeth are in occlusion, resulting in 555 teeth available for the FDI-labeling task.

\begin{figure}[h]
    \centering
    \includegraphics[width=0.7\linewidth, trim=0cm 25cm 0cm 26cm, clip]{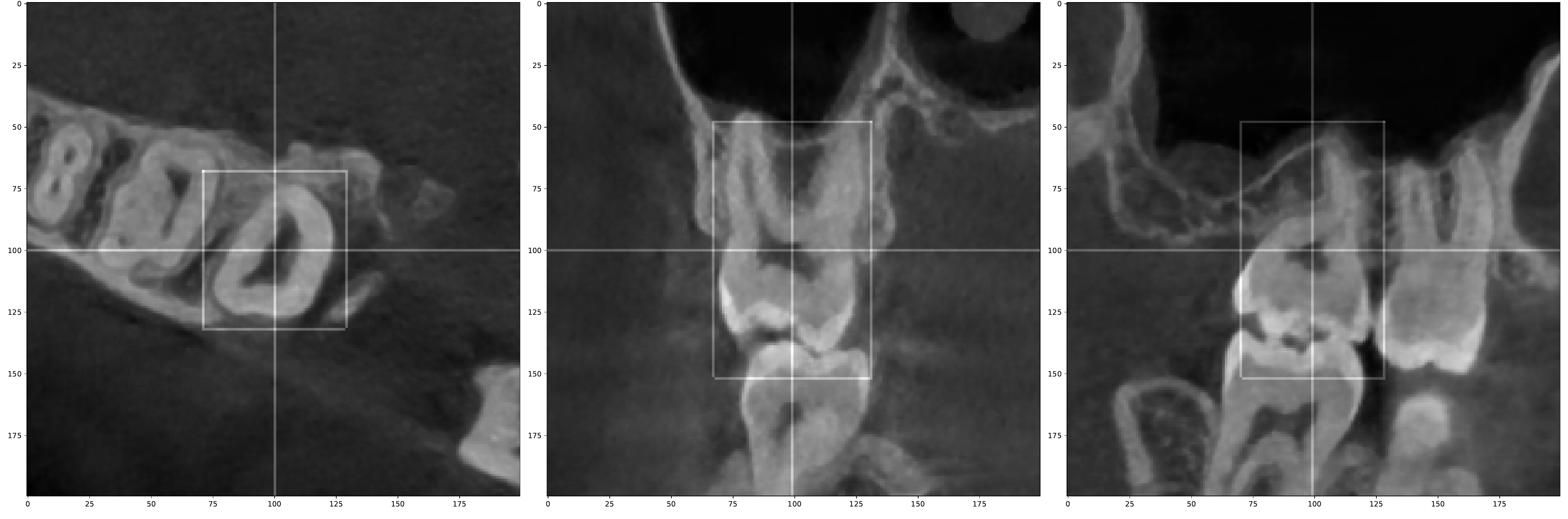}
    \caption{{Three two dimensional slices through a CBCT-scan, tooth 16 is depicted within its ROI and surrounding structures are included.}}
    \label{fig:ROI}
\end{figure}

\subsubsection{Study Design}
Our proposed \emph{pipeline} to classify the CBCT data is as follows:

\begin{enumerate}

\item {\bfseries Generating ROI from CBCT-data}
We generated regions of interest (ROIs) containing individual teeth, as seen in Figure \ref{fig:ROI}, via coarse segmentation of dental structures (teeth, maxilla, mandible) using a deep learning approach. For this step, a 3D U-Net, implemented in MONAI (v1.3) with PyTorch (v2.1), was trained on 120 CBCT datasets with 15 for validation. Standard preprocessing and 3D data augmentation were applied. The model was trained with the Adam optimizer and a Dice–Cross-Entropy loss. After training, the network was used to segment new CBCT volumes (NIfTI) \citep{Anonymous2022}. From each coarse tooth segmentation, a 3D bounding box was computed, expanded by 3 voxels in all directions to ensure full coverage, and the corresponding sub-volumes (ROIs) were extracted from the CBCT data.  For more information, see Appendix \ref{The size and location of the ROI}.
\item {\bfseries Constructing Filtrations}
For classification, we use the full three-dimensional ROI together with forty-one two-dimensional slices in the coronal, sagittal, and axial directions.  For details, see Appendix \ref{The location of the two dimensional slices}. We do not perform any normalization of the grayscale values, as this outperforms all normalization attempts (see Appendix \ref{The used normalization method}).
From the three-dimensional ROI and the slices, we consider the {\bfseries original}, {\bfseries inverted}, {\bfseries contrast}, and {\bfseries inverted contrast images} and construct the corresponding sublevel set filtrations as described in Subsection~\ref{subsec:cubical_filtration}.

\item {\bfseries Computing Persistence Diagrams}
From the filtrations obtained in Step 2 we construct persistence diagrams in each dimension less or equal to the one of the image or volume respectively, resulting in $(4\cdot3 )+((41\cdot3)\cdot4\cdot2)=996$ persistence diagrams in total per ROI.

\item {\bfseries Vectorization}
To assign vectors to persistence diagrams, we use the code provided in \citep{Ali_2023}, specifically {\bfseries persistent statistics}. Doing so, we extract 38 features out of each persistence diagram (mean or quantiles of births or deaths, etc.). For the results obtained with the other vectorization techniques, see Appendix~\ref{The used vectorization method}.

%, having 38 features. %\textcolor{red}{ In \citep{pmid30484214} it was shown that texture features can improve caries diagnostics, we therefore append texture features from the three-dimensional ROI as well as from all forty-one two-dimensional slices.} The features include the mean, variance, median, mode, skewness, kurtosis, energy, entropy, minimum, maximum, contrast ratio, 10th percentile intensity, 25th percentile intensity, 75th percentile intensity, 90th percentile intensity, and the interpercentile range of the grayscale values.
\item {\bfseries SVM}
Through persistent statistics  we generate 37848 features. % and an additional 1985 from texture.
In order to reduce the number of features, we train a linear SVM with $C = 0.001$ and order the features by their assigned weights. We then only use the features with the highest weights to perform 5-fold cross-validation, iteratively increasing the number of used features. We keep track of the highest accuracy; once the score does not improve by adding more features, we stop. Appendix \ref{The SVM structure} summarizes additional experiments, including comparisons of SVM kernels, the incorporation of texture-analysis features, and several manual feature-selection strategies, none of which improved the test accuracy.

\item{\bfseries Iterated labeling}
In order to decrease the number of labels that need to be differentiated at once, we perform iterative steps, each narrowing down the final label. The individual steps are described in Figure \ref{flowcharts}.

\end{enumerate}

\begin{figure}
\floatconts
  {flowcharts}
  {\caption{Flowcharts for performing the FDI-labeling and diagnosis tasks.}}
  {%
    \subfigure[Flowchart for performing the FDI-labeling task. (In detail for the first quadrant.)]{\label{flowchart:FDI}%
      \includegraphics[page=3, scale=0.75, clip, trim=0cm 8cm 0cm 0cm, width=0.48\textwidth]{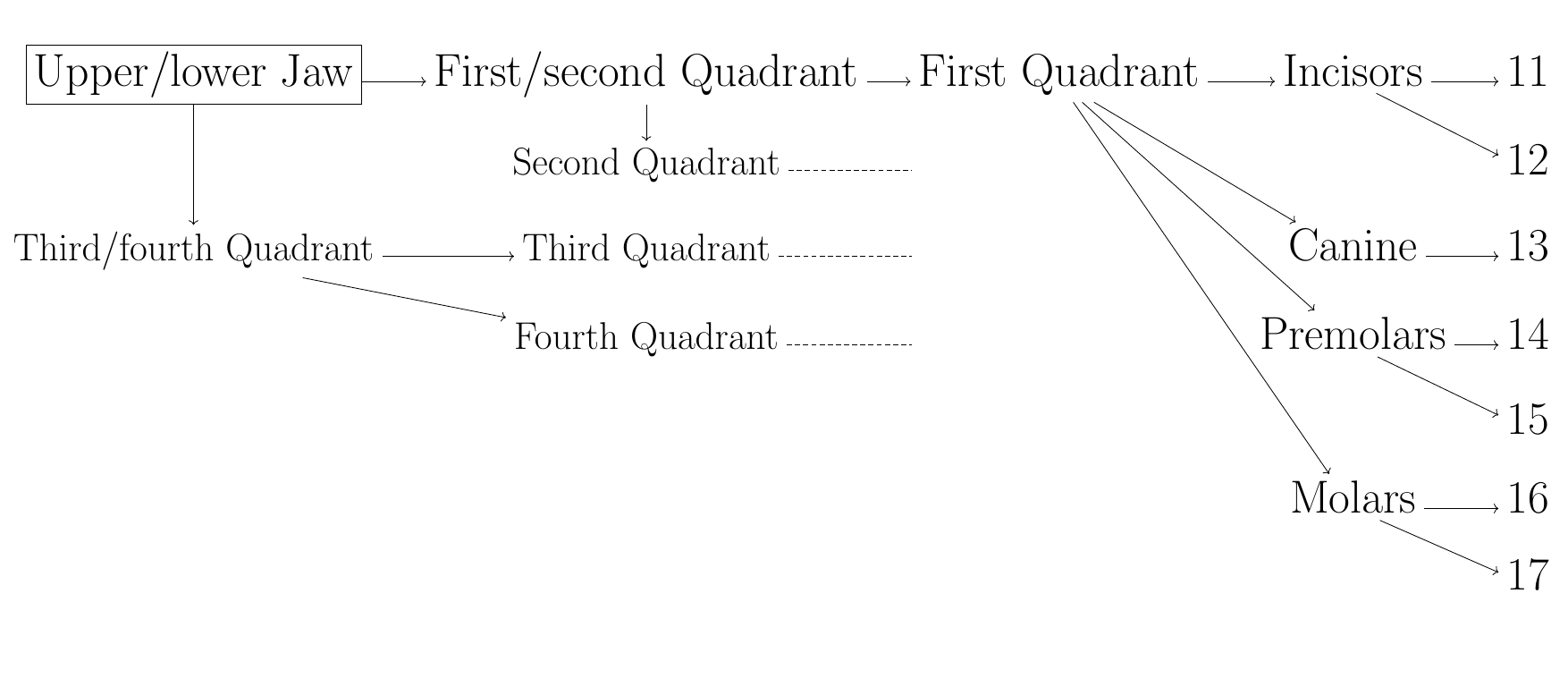}}%
    %\qquad
    \hfill
    \subfigure[Flowchart for performing the diagnosis task.]{\label{flowchart:diagnosis}%
      \includegraphics[page=6, scale=0.75, clip, trim=0cm 8.8cm 0cm 0cm, width=0.48\textwidth]{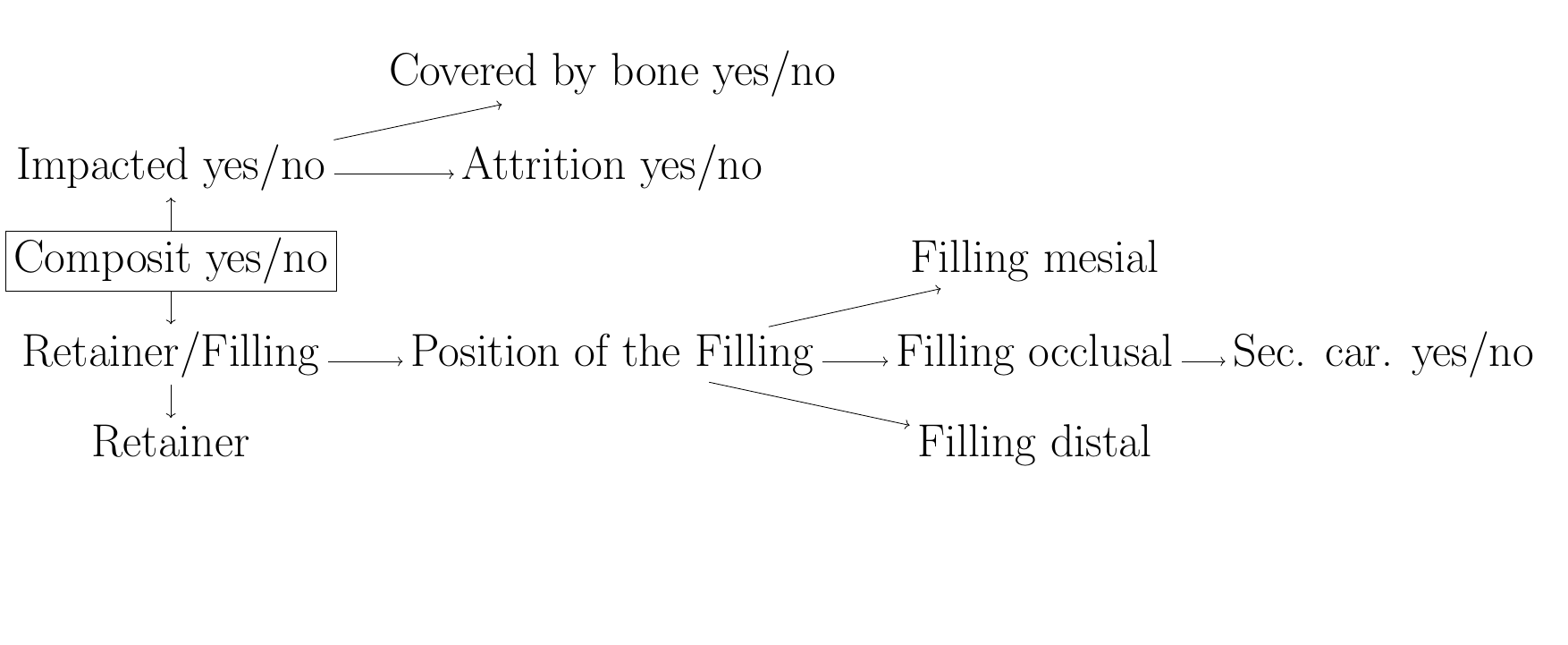}}
  }
\end{figure}

To perform the FDI-labeling, we first differentiate between the upper and lower jaws, then identify the quadrant, followed by determining whether the tooth is an incisor, canine, premolar or molar, and finally assign the appropriate label. This can be seen in Figure \ref{flowchart:FDI}. Note that this allows us to first have binary decision then a quaternary decision and finaly again a binary, or none at all if we have a canine. Additionally, this allows us to mirror the teeth of the second and third quadrant along the sagittal plane for the final two steps, in order to double the amount of samples we have in each quadrant.

% \begin{figure}
%     \centering
%     \subcaptionbox{Flowchart for performing the FDI-labeling task.\label{flowchart:FDI}}{%
%         \includegraphics[page=3, scale=0.75, clip, trim=0cm 8cm 0cm 0cm, width=0.48\textwidth]{Images/flowcharts2.pdf}
%     }
%     \hfill
%     \subcaptionbox{Flowchart for performing the diagnosis task.\label{flowchart:diagnosis}}{%
%         \includegraphics[page=6, scale=0.75, clip, trim=0cm 8.8cm 0cm 0cm, width=0.48\textwidth]{Images/flowcharts2.pdf}
%     }
%     \caption{Flowcharts for performing the FDI-labeling and diagnosis tasks.}
% \label{flowcharts}
% \end{figure}

To perform a diagnosis, we first check whether composite material is present in the region of interest. If not, we determine whether the tooth is impacted. If it is, we differentiate between teeth covered by bone and those covered only by gingiva. Otherwise, we assess for attrition damage. If composite material is present in the region of interest, we begin by distinguishing between retainers and fillings. For fillings, we also evaluate their position and check whether the tooth is affected by secondary caries. This can be seen in Figure \ref{flowchart:diagnosis}. %The texture features, while only improving the accuracy marginal if at all, helped reduce the number of feature needed to reach said accuracy.

In both tasks, we initially start with the whole dataset, meaning $n=555$ for the FDI-labeling task and $n=600$ in order to perform diagnostics. Each step reduces the dataset, since teeth excluded at earlier stages are removed from subsequent classification steps.
%Each step reduces the data since for the subsequent labeling task, as all teeth that can be excluded are removed from the dataset. 
For example, in the decision step whether the given tooth is in the first or second quadrant we only use teeth from the upper jaw, as the previous step already differentiated between upper and lower jaw. The only exception is the step in which we differentiate between incisor, canine, premolar and molar, as we mirrored the teeth from the other quadrant in order to double our sample size. In each step, we use 5-fold cross-validation to split the data in training and testing.\footnote{We provide our code at https://github.com/leon-dahlmeier/Topology-of-a-smile}

\section{Results}

The scores and number of features used for each individual decision step can be seen in Tables \ref{Table:FDI-overall} and \ref{Table:diagnosis-labels}. The resulting scores for each label can be seen in Table \ref{Table:iterated-scores}. In the FDI-labeling task, this results in scores ranging from 96.60\% to 98.57\% averaging 97.67\% for all labels. This corresponds to the average of 96.78\% in the first/second quadrant and 98.57\% in the third/fourth quadrant.
%In the first/second quadrant, this averages to 96.78\% and 98.57\% in the third/fourth quadrant. 
For performing diagnostics, this reaches scores from 88.41\% to 98.17\%, with an average of 96.77\%. We have also provided some additional metrics, such as AUC, recall, and precision, where it makes sense.

\begin{table}[t]
%\centering
\floatconts
 {Table:FDI-overall}
 {\caption{Overview of classification scores. The first table shows quadrant differentiation, while the second table compares tooth classification scores in the upper and lower jaws: accuracy (in percentages), AUC in brackets and the number of features in parentheses.}}
 {%
   \subtable[Scores reached for differentiating between quadrants.]{%
     \label{Table:FDI-first-labels}%
     \begin{tabular}{|l|c|c|c|}
        \hline
        \textbf{Labels} & \textbf{ Accuracy (\%) } & \textbf{AUC } & \textbf{ \# of Features } \\ \hline \hline
        Upper/lower Jaw & 100 & 1 & 111 \\ \hline \hline
        First/second quadrant & 98.55 &0.9892 & {1411} \\ \hline
        Third/fourth quadrant & 98.93 &0.9839 & {911} \\ \hline
    \end{tabular}
   }\qquad
   \subtable[Scores for tooth classification in the upper and lower jaws.]{%
     \label{Table:FDI-upper-lower-jaws}%
     \begin{tabular}{|l|c|c|}
        \hline
        \textbf{} & \textbf{ Upper Jaw } & \textbf{Lower Jaw} \\ \hline
        \textbf{Labels} & \multicolumn{2}{c|}{\textbf{Acc. (\%) [AUC] (\# of Features)}} \\ \hline \hline
        Types of teeth & 99.27 [-] (1394) & 99.64 [-] (1084) \\ \hline
        First / second incisor & 98.75 [1] {(111)} & 100 [1]{(811)} \\ \hline
        Canine & 100 [1] (0) & 100 [1] (0) \\ \hline
        First / second premolar & 98.75 [0.975] {(311)} & 100 [1] (611) \\ \hline
        First / second molar & 98.75 [0.9906] {(511)} & 100 [1] {(2211)} \\ \hline
    \end{tabular}
   }
 }
\end{table}

To compare these results to a convolutional neural network (CNN), we additionally trained a CNN on our dataset. We chose not to use proprietary commercial software as a benchmark, as its substantially larger training datasets and undisclosed architectures and training procedures preclude reproducible comparisons under conditions comparable to ours. Instead, we adopted the input configuration proposed by \citet{ESMAEILYFARD2024328}, using one image from each anatomical plane (axial, sagittal, and coronal) as input. Given that this architecture achieved competitive results in their study and that we began to observe signs of overfitting, we decided not to include additional slices in the CNN training process. The network consists of two convolutional blocks, each comprising a $3\times3$ convolution with 32 feature maps, followed by a ReLU activation and a $2\times2$ max pooling layer. The extracted features are flattened and passed through a fully-connected layer with 128 units and ReLU activation, with dropout applied for regularization, before a final linear layer outputs class probabilities. Input images were normalized and augmented during training with random resized cropping, Gaussian noise, and random erasing. The model was trained using the Adam optimizer with a learning rate of $10^{-3}$ for 500 epochs and achieved test accuracies of 70.27\% for FDI labeling and 86.67\% for diagnostics. Appendix~\ref{The choice of CNN} summarizes the selection of the CNN baseline and the alternatives we considered.
%test classification accuracy of 70.27\% for FDI-labels and 86.67\% for diagnostics.

\begin{table}[t]
\caption{Scores reached in the diagnosis task.}
\label{Table:diagnosis-labels}
\centering
\begin{tabular}{|l|r|r|r|r|}
\hline
\textbf{Labels} & \textbf{Accuracy (\%)} &\textbf{recall} & \textbf{precision} & \textbf{\# of Features} \\ \hline \hline
Composite yes/no & 98.17 & 0.97 &  0.95 & {1811} \\ \hline \hline
Impacted yes/no & 100 & 1 & 1 & 711 \\ \hline
Covered by bone yes/no & 100 & 1 & 1 & 711 \\ \hline
Attrition/nothing & 99.58 & 0.67 & 0.67 & 1211 \\ \hline \hline
Filling/Retainer & 100 & 1 & 1 & {211} \\ \hline
Position of the filling &  96.00 &-&-& {464} \\ \hline
Secondary caries & {95.56} & 0.8 & 1 & 111 \\ \hline
%Secondary caries (precision) & \textcolor{red}{80} & 211 \\ \hline
\end{tabular}
\end{table}

\begin{table}[!ht]
\centering
\caption{Iterated scores for all final predictions, together with overall averages.}
\label{Table:iterated-scores}
\begin{minipage}{0.4\textwidth}
    \centering
    \begin{tabular}{|l|c|}
    \hline
    \textbf{FDI-Labels} & \textbf{Acc. (\%)} \\ \hline \hline
    12, 11, 21, 22 & 96.60  \\ \hline
    13, 23 & 97.83 \\ \hline
    14, 15, 24, 25 & 96.60\\ \hline
    16, 17, 26, 27 & 96.60\\ \hline \hline
    42, 41, 31, 32 & 98.57  \\ \hline
    43, 33 & 98.57 \\ \hline
    44, 45, 34, 35 & 98.57\\ \hline
    46, 47, 36, 37 & 98.57\\ \hline
    \end{tabular}
\end{minipage}
\hfill
\begin{minipage}{0.59\textwidth}
    \centering
    \begin{tabular}{|l|c|c|c|}
    \hline
    \textbf{Diagnostic Labels} & \textbf{Acc. (\%)}& \textbf{recall }&\textbf{precision }\\ \hline \hline
    Impacted & 98.17&0.97&0.95\\ \hline
    Covered by bone & 98.17&0.97&0.95\\ \hline
    Attrition & 97.75& 0.67&0.67\\ \hline
    No findings & 97.75&0.97& 0.95\\ \hline
    Filling (with pos.) & 94.24&0.98 &0.94\\ \hline
    Secondary caries & 88.44& 0.8 &0.94\\ \hline
    Retainer & 98.17 & 0.97&0.95\\ \hline
    \end{tabular}
\end{minipage}
\vspace{0.5cm}\\
\begin{tabular}{ |c|c|c| } 
     \hline
     & \textbf{Our methods (\%)} & \textbf{CNN reference method (\%)}  \\ \hline
     FDI-Labels & 97.67 & 70.27   \\ \hline
     Diagnostic & 96.77 & 86.67   \\ \hline
    \end{tabular}
\end{table}

\section{Conclusion}

We introduce an automated pipeline for tooth classification and diagnosis. Regions of interest (ROIs) are extracted from CBCT data using deep learning methods, while classification is performed via persistent homology. On datasets comparable to those reported in the literature, we reach average accuracy scores of 97.67\% for tooth labeling and 96.77\% for diagnostic tasks, outperforming a CNN trained on the same data with accuracy of 70.27\% and 86.67\% respectively.

%The labeling and diagnosing in our workflow is grounded in a principled multi-scale theory and achieves performance comparable to the literature on similar datasets, reaching average accuracy scores of 97.74\% for tooth-labeling and 96.10\% for diagnostics tasks, outperforming a CNN trained on the same data with accuracy of 70.27\% and 86.67\% respectively.
%\\
The labeling and diagnosing in our workflow relies on small datasets and few features, enabling faster and less computationally demanding training. It also accommodates regions of interest of varying size. 

%Finally, our model is \emph{robust}: unlike CNNs, which can be sensitive to small voxel-level perturbations, our approach benefits from the provable stability of persistent homology~\citep{Cohen-Steiner2007,kim2021noise}.

By automating tooth classification, our approach reduces clinician workload and supports more efficient diagnostics. Future work includes increasing the sample size to further improve performance and extending the persistent homology framework to tooth detection itself, thereby eliminating the need for a separate segmentation step.

\clearpage

\acks{
    This work has received funding from the Swiss State Secretariat for
Education, Research, and Innovation~(SERI).
}

\bibliography{main.bib}

\clearpage

\appendix

\section{First Appendix}\label{apd:first}

\newcommand{\shadeFirstTable}[1]{
    \ifdim #1 pt < 76 pt \cellcolor{red!15}\else
    \ifdim #1 pt < 78 pt \cellcolor{orange!20}\else
    \ifdim #1 pt < 80 pt \cellcolor{yellow!20}\else
    \ifdim #1 pt < 82 pt \cellcolor{yellow!30}\else
    \ifdim #1 pt < 84 pt \cellcolor{green!30}\else
    \ifdim #1 pt < 86 pt \cellcolor{green!40}\else
    \ifdim #1 pt < 88 pt \cellcolor{green!50}\else
    \cellcolor{green!60}\fi\fi\fi\fi\fi\fi\fi
}

% Define the shade command for the second table (broader gradation)
\newcommand{\shadeSecondTable}[1]{
    \ifdim #1 pt < 50 pt \cellcolor{red!30}\else
    \ifdim #1 pt < 70 pt \cellcolor{orange!30}\else
    \ifdim #1 pt < 80 pt \cellcolor{yellow!30}\else
    \ifdim #1 pt < 90 pt \cellcolor{green!30}\else
    \cellcolor{green!50}\fi\fi\fi\fi
}

We will now elaborate on the ablation studies we performed and the decisions we took, in order to settle on 
\ref{The used normalization method} the normalization method,
\ref{The size and location of the ROI} the tooth detection and its impact on the ROI, 
\ref{The location of the two dimensional slices} the location of two dimensional slices,
\ref{The used vectorization method} the vectorization method,
\ref{The SVM structure} the SVM structure, 
\ref{The choice of CNN} the choice of CNN.
Since the methodology provided in this section is not always the final approach, all scores should be understood relatively within the different experimentation rather than absolutely.
\\We started by using the three-dimensional voxel set of the whole tooth, together with five two-dimensional slices in each anatomical direction. From there we used the filtration as seen in Subsubsection \ref{subsec:cubical_filtration}.

\subsection{Normalization method\label{The used normalization method}:}
  Since the images analyzed in this manuscript were not all acquired using the same machine, the ranges of grayscale values vary substantially between images. A common way to account for this variability when using SVM would be to normalize the images before classification. In Table \ref{table:normalization} we see the results of this. We normalized either the whole X-ray or just the region of interest in which the specified tooth lies. In each case, the grayscale values were rescaled to the intervals $[0,1]$, $[0,10,000]$, and $[0,30,000]$.
    
    The interval $[0,1]$ was included because it is the most commonly used normalization range. The range $[0,10,000]$ was chosen to investigate the effect of preserving a larger portion of the original grayscale variation, as the dynamic range differs substantially between images. Finally, the range $[0,30,000]$ was selected because it exceeds the grayscale range observed in any image of the dataset, thereby providing a setting with minimal compression of intensity values.
    
    Overall, the highest classification accuracy was achieved without any normalization. The way the filtrations are set up, there is exactly one feature persisting until infinity, specifically the first connected component. Since the vectorization methods cannot work with the value `inf' we exchange it for the highest gray-scale-value $+ 1$ in a given image. This choice ensures that the feature still records the maximal grayscale value of the image while remaining finite, allowing it to be processed by the vectorization methods and subsequently by the SVM. 

\begin{table}[ht!]
\centering
\begin{tabular}{|l|c|c|}
\hline
\textbf{Normalization Method} & \textbf{Testing Accuracy} (\%)& \textbf{Training Accuracy} (\%) \\\hline
\hline
whole image to $[0,1]$ & \shadeFirstTable{57.29}57.29 & 100.00 \\\hline
tooth to $[0,1]$ & \shadeFirstTable{65.95}65.95 & 100.00 \\\hline
whole image to $[0,10'000]$ & \shadeFirstTable{86.67}86.67 & 100.00 \\\hline
tooth to $[0,10'000]$ & \shadeFirstTable{86.49}86.49 & 100.00 \\\hline
whole image to $[0,30'000]$ & \shadeFirstTable{86.31}86.31 & 100.00 \\\hline
tooth to $[0,30'000]$ & \shadeFirstTable{86.18}86.18 & 100.00 \\\hline
no normalization & \shadeFirstTable{88.29}88.29 & 100.00 \\\hline
\end{tabular}
\caption{Training and testing accuracy for various normalization methods with Testing accuracy heat map.}
\label{table:normalization}
\end{table}

\subsection{Tooth detection and impact on ROI\label{The size and location of the ROI}:}
  
    This aspect warrants further investigation. In the present work, we adopted the segmentation-based approach in order to obtain a fully automatic workflow.
    
    The exact impact of suboptimal selected ROI's still needs to be further investigated. But we have run our tests twice, once using the segmentation approach described above and once using manually selected ROI's, ensuring voxel level precision.
    \begin{itemize}
        \item To reiterate the results when using the \textbf{segmentation approach}: In the FDI-labeling task, this results in scores ranging from 96.60\% to 98.57\% averaging 97.67\% for all labels. For performing diagnostics, this reaches scores from 88.41\% to 98.17\%, with an average of 96.77\%.
        \item And in comparison the \textbf{manually selected} ROI's :For the FDI-labeling task, we reached scores ranging from 97.31\% to 98.55\% averaging at 97.89\%. For performing diagnostics, we reached scores from 90.96\% to 99.13\% averaging at 96.77\%. 
    \end{itemize}

\subsection{Slice location\label{The location of the two dimensional slices}:}

  We further investigated the impact of changing the locations of the two-dimensional slices. To this end, we fixed one slice at the midpoint of each dimension and varied the positions of the remaining four slices. Using the parameters `inner' and `outer', we decided the position of the remaining slices as follows: the two slices closest to the slice in the middle were placed at position ${(10\pm \text{inner})\cdot(\text{dimension}/20)}$, while the outermost slices were placed at ${(\text{outer})\cdot(\text{dimension}/20)}$ and ${(20-\text{outer})\cdot(\text{dimension}/20)}$. The resulting performance is reported in Table~\ref{table:slice-location}. We adopted the best-performing configuration for all subsequent experiments.

\begin{table}[ht!]
    \centering
    \begin{tabular}{|l||*{5}{p{3.5em}|}}\hline
    \backslashbox{outer}{inner}
    &\makebox[3em]{1}&\makebox[3em]{2}&\makebox[3em]{3}&\makebox[3em]{4}&\makebox[3em]{5}\\\hline\hline
    1 & \shadeFirstTable{78.38}78.38\% & \shadeFirstTable{83.78}83.78\% & \shadeFirstTable{82.88}82.88\% & \shadeFirstTable{84.68}84.68\% & \shadeFirstTable{82.88}82.88\% \\\hline
    2 & \shadeFirstTable{81.08}81.08\% & \shadeFirstTable{82.88}82.88\% & \shadeFirstTable{83.78}83.78\% & \shadeFirstTable{87.39}87.39\% & \shadeFirstTable{84.68}84.68\% \\\hline
    3 & \shadeFirstTable{81.08}81.08\% & \shadeFirstTable{86.49}86.49\% & \shadeFirstTable{88.29}88.29\% & \shadeFirstTable{85.59}85.59\% & \shadeFirstTable{83.78}83.78\% \\\hline
    4 & \shadeFirstTable{77.48}77.48\% & \shadeFirstTable{82.88}82.88\% & \shadeFirstTable{83.78}83.78\% & \shadeFirstTable{80.18}80.18\% & \shadeFirstTable{82.88}82.88\% \\\hline
    5 & \shadeFirstTable{75.68}75.68\% & \shadeFirstTable{83.78}83.78\% & \shadeFirstTable{85.59}85.59\% & \shadeFirstTable{83.78}83.78\% & \shadeFirstTable{81.98}81.98\% \\\hline
    \end{tabular}
    \caption{Different locations for the two-dimensional slices with heat map.}
    \label{table:slice-location}
\end{table}

\subsection{Vectorization method\label{The used vectorization method}:}
Using the code provided by \citep{Ali_2023}, we compared a range of vectorization methods to determine the one best suited for our task. Descriptions of the methods not discussed in Subsubsection~\ref{Subsubsection:Vectorization Techniques} can be found in that reference.
 In Table \ref{table:vectorization} we can see that extracting statistical features generated the highest accuracy. We therefore adopted this vectorization method for all subsequent experiments.

\begin{table}[ht!]
\centering
\begin{tabular}{|l|c|c|}
\hline
\textbf{Vectorization Method} & \textbf{Testing Accuracy} (\%)& \textbf{Training Accuracy} (\%) \\\hline
\hline
PersStats & \shadeSecondTable{88.29}88.29 & 100.00 \\\hline
AlgebraicFunctions & \shadeSecondTable{75.68}75.68 & 100.00 \\\hline
BettiCurveFeature & \shadeSecondTable{71.17}71.17 & 100.00 \\\hline
ComplexPolynomialFeature & \shadeSecondTable{74.77}74.77 & 100.00 \\\hline
PersImageFeature & \shadeSecondTable{34.35}34.35 & 100.00 \\\hline
PersLandscapeFeature & \shadeSecondTable{87.39}87.39 & 100.00 \\\hline
PersLifespanFeature & \shadeSecondTable{78.38}78.38 & 100.00 \\\hline
PersSilhouetteFeature & \shadeSecondTable{76.58}76.58 & 100.00 \\\hline
PersTropicalCoordinatesFeature & \shadeSecondTable{82.88}82.88 & 100.00 \\\hline
\end{tabular}
\caption{Training and testing accuracy for various vectorization methods with testing accuracy heat map.}
\label{table:vectorization}
\end{table}

\subsection{The SVM structure:}\label{The SVM structure}

    So far we used a grid search to find the best kernel for the SVM, including polynomial, radial basis function (RBF), and sigmoid kernels. A linear kernel which $C=0.001$ consistently performed the best. We therefore settled on skipping the grid search for all following iterations, in order to combat overfitting.

    Additional features such as Texture analysis did not improve test accuracy.

    The tests described up until now including this one the three-dimensional voxel set of the whole ROI, together with five two-dimensional slices in each anatomical direction. In order to combat overfitting, as seen in Table \ref{table:vectorization} and to understand which features are the most important, we reduced the number of filtrations. In Table \ref{table:reduce_to_accuracy} we see on the left what remains, that is, whether or not the three-dimensional image was used, and from the two-dimensional image we differentiate between the coronal sagital and axial slices as well as the homology of dimension one and zero. We ran these tests for both the statistical and the landscape-feature approach to vectorization. We primarily focused on reducing to the two-dimensional images, from which we also either extracted only one- or zero dimensional homology. We expected dimension 1 homology to be the more important feature compared to dimension 0. We therefore additionally started by only using dimension one homology of the two-dimensional images, but further left out the three-dimensional images or the two-dimensional slices with either even or odd index. We want to point out the row using all features previously used, resulting in the best accuracy. Leaving out the three-dimensional region of interest only slightly reduced the overall accuracy. 

    We therefore settled on not manually reducing our features but rather using a more methodical approach, akin to principal component analysis (PCA). Specifically we train a linear SVM with $C = 0.001$ and order the features by their assigned weights. We then only use the features with the highest weights to perform 5-fold cross-validation, iteratively increasing the number of used features.
    
\begin{table}[ht!]
\centering

% \begin{tabular}{|l|c|c|c|}
% \hline
% \textbf{Reduced to} & \textbf{Stat. test acc. (\%)} & \textbf{Ladsc. test acc. (\%)} & \textbf{Train acc. of both (\%)} \\
% \hline \hline
% coronal & \shadeFirstTable{74.77}74.77 & \shadeFirstTable{64.32}64.32 & 100.00 \\ \hline
% sagittal & \shadeFirstTable{75.14}75.14 & \shadeFirstTable{64.14}64.14 & 100.00 \\ \hline
% d1 (even) & \shadeFirstTable{75.32}75.32 & \shadeFirstTable{69.91}69.91 & 100.00 \\ \hline
% axial & \shadeFirstTable{75.86}75.86 & \shadeFirstTable{75.32}75.32 & 100.00 \\ \hline
% d1 (odd) & \shadeFirstTable{80.54}80.54 & \shadeFirstTable{73.33}73.33 & 100.00 \\ \hline
% d1 & \shadeFirstTable{86.13}86.13 & \shadeFirstTable{78.92}78.92 & 100.00 \\ \hline
% d1 (no 3D) & \shadeFirstTable{86.31}86.31 & \shadeFirstTable{77.66}77.66 & 100.00 \\ \hline
% everything & \shadeFirstTable{87.57}87.57 & \shadeFirstTable{81.26}81.26 & 100.00 \\ \hline
% d0 & \shadeFirstTable{87.75}87.75 & \shadeFirstTable{74.41}74.41 & 100.00 \\ \hline
% \end{tabular}

\begin{tabular}{|c|c|c|c|c|c|c|c|c|}
\hline
\begin{sideways}\textbf{3D}\end{sideways} 
& \begin{sideways}\textbf{d1 coronal}\end{sideways} 
& \begin{sideways}\textbf{d1 sagital}\end{sideways} 
& \begin{sideways}\textbf{d1 axial}\end{sideways} 
& \begin{sideways}\textbf{d0 coronal}\end{sideways} 
& \begin{sideways}\textbf{d0 sagital}\end{sideways} 
& \begin{sideways}\textbf{d0 axial}\end{sideways} 
& \begin{sideways}\textbf{Stat. acc. (\%)}\end{sideways}
& \begin{sideways}\textbf{Ladsc. acc. (\%)}\end{sideways} \\
\hline \hline

\checkmark & \checkmark & & & \checkmark & &
& \shadeFirstTable{74.77}74.77
& \shadeFirstTable{64.32}64.32 \\ \hline

\checkmark & & \checkmark & & & \checkmark &
& \shadeFirstTable{75.14}75.14
& \shadeFirstTable{64.14}64.14 \\ \hline

\checkmark & & & \checkmark & & & \checkmark
& \shadeFirstTable{75.86}75.86
& \shadeFirstTable{75.32}75.32 \\ \hline

\checkmark & \checkmark & \checkmark & \checkmark & & &
& \shadeFirstTable{86.13}86.13
& \shadeFirstTable{78.92}78.92 \\ \hline

& \checkmark & \checkmark & \checkmark & & &
& \shadeFirstTable{86.31}86.31
& \shadeFirstTable{77.66}77.66 \\ \hline

\checkmark & \checkmark & \checkmark & \checkmark & \checkmark & \checkmark & \checkmark
& \shadeFirstTable{87.57}87.57
& \shadeFirstTable{81.26}81.26 \\ \hline

\checkmark & & & & \checkmark & \checkmark & \checkmark
& \shadeFirstTable{87.75}87.75
& \shadeFirstTable{74.41}74.41 \\ \hline

\end{tabular}

\caption{Test and train accuracy scores for fewer two-dimensional images with heat map shading.}
\label{table:reduce_to_accuracy}
\end{table}

\subsection{The choice of CNN:}\label{The choice of CNN}
We decided against using proprietary commercial software as a baseline, since such systems are typically trained on substantially larger datasets and their architectures and training procedures are not publicly available. As a result, they cannot be reproduced or trained under conditions comparable to our own.

We therefore considered two different approaches:
\begin{enumerate}
    
    \item Using the pre-trained ResNet50, we generate vectors for three axial slices of the tooth. These vectors were concatenated and used as data to train and test an MLP resulting in an accuracy of 82.88\% for FDI labeling and 78.87\% for diagnostics. Unfortunately, ResNet50 is trained on a bigger dataset than what we worked with and is not trained specifically for teeth making it harder to compare with our approach.

    \item We settled on using the approach described in \citep{ESMAEILYFARD2024328}. The advantages are twofold:
    \begin{itemize}
       \item The architecture is publicly available, allowing us to implement and train the network ourselves rather than relying on a proprietary black-box system.
\item The network is trained exclusively on our dataset, ensuring a fair comparison with our topological approach, since both methods use the same training data.
    \end{itemize}
\end{enumerate}

%\section{Second Appendix}\label{apd:second}

%This is the second appendix.

\end{document}